%% file: emnlp2021.tex
\pdfoutput=1

\documentclass[11pt]{article}

\usepackage{emnlp2021}

\usepackage{times}
\usepackage{latexsym}

\usepackage[T1]{fontenc}

\usepackage[utf8]{inputenc}

\usepackage{microtype}

\usepackage{graphicx}
\usepackage{subcaption}

\usepackage{booktabs, tabularx}

\usepackage{todonotes}

\title{Language Modeling, Lexical Translation, Reordering:\\ The Training Process of NMT through the Lens of Classical SMT}

  \author{Elena Voita$^{1,2}$\quad\quad Rico Sennrich$^{3,1}$\quad\quad Ivan Titov$^{1,2}$\bigskip\\
  $^1$University of Edinburgh, Scotland  \quad
  $^2$University of Amsterdam, Netherlands \\
  $^3$University of Zurich, Switzerland \\
  {\tt lena-voita@hotmail.com} \quad {\tt sennrich@cl.uzh.ch} \quad {\tt ititov@inf.ed.ac.uk}
}

\begin{document}
\maketitle
\begin{abstract}

Differently from the traditional statistical MT that decomposes the translation task into distinct separately learned components, neural machine translation uses a single neural network to model the entire translation process. Despite neural machine translation being de-facto standard, it is still not clear how NMT models acquire different competences over the course of training, and how this mirrors the different models in traditional SMT. In this work, we look at the competences related to three core SMT components and find that during training, NMT first focuses on learning target-side language modeling, then improves translation quality approaching word-by-word translation, and finally learns more complicated reordering patterns. We show that this behavior holds for several models and language pairs. Additionally, we explain how such an understanding of the training process can be useful in practice and, as an example, show how it can be used to improve vanilla non-autoregressive neural machine translation by guiding teacher model selection.

\end{abstract}

\section{Introduction}

In the last couple of decades, the two main machine translation paradigms have been statistical and neural MT. Statistical MT (SMT) decomposes the translation task into several components (e.g., lexical translation probabilities, alignment probabilities, target-side language model, etc.) which are learned separately and then combined in a translation model. Differently, neural MT (NMT) models the entire translation process with a single neural network that is trained end-to-end.

Although joint training of all the components is one of the obvious NMT strengths, this is also one of its challenging aspects. While SMT models different competences with distinct model components and, therefore, can easily validate and/or improve each of them, NMT acquires these competences within the same network over the course of training. Even though previous work shows how to improve some of the competences in NMT, e.g., by using lexical translation probabilities, phrase memories,  target-side LM, alignment information~(\citealp{arthur-etal-2016-incorporating,He2016ImprovedNM,tang2016neural,wang-etal-2017-translating,zhang-etal-2017-prior,dahlmann-etal-2017-neural,Gulcehre2015,Glehre2017OnIA,He2016ImprovedNM,Sriram2017ColdFusion,dahlmann-etal-2017-neural,stahlberg-etal-2018-simple,mi-etal-2016-supervised,liu-etal-2016-neural,chen2016guided,alkhouli-etal-2016-alignment,alkhouli-ney-2017-biasing,park-tsvetkov-2019-learning,Song2020AlignmentEnhancedTF} among others), it is still not clear how and when NMT acquires these competences during training. For example, are there any stages where NMT focuses on different aspects of translation, e.g., fluency (agreement on the target side) or adequacy (i.e.\ connection to the source), or does it improve everything at the same rate? Does it learn word-by-word translation first and more complicated patterns later, or is there a different behavior? This is especially interesting in light of a recent work analyzing how NMT balances the two different types of context: the source and prefix of the target sentence~\cite{voita2021analyzing}. As it turns out, changes in NMT training are non-monotonic and form several distinct stages (e.g., stages changing direction from decreasing influence of source to increasing), which hints that the NMT training consists of stages with qualitatively different changes.

In this paper, we try to understand what happens in these stages by analyzing translations generated at different training steps. Specifically, we focus on the aspects related to the three core SMT components: target-side language modeling, lexical translation, and reordering. We find that during training, NMT focuses on these aspects in the specified above order.
Intuitively, it starts by hallucinating frequent n-grams and sentences in the target language, then comes close to word-by-word translation, and finally learns more complicated reordering patterns. We confirm these findings for several models, LSTM and Transformer, and different modeling paradigms, encoder-decoder and decoder-only, i.e.\ LM-style machine translation where a left-to-right language model is trained on the concatenation of source and target sentences.

Finally, we show how such an understanding of the training process can be useful in practice. Namely, we note that during a large part of training, a model's quality (e.g. BLEU and token-level predictive accuracy) changes little, but reordering becomes more complicated. This means that by using different training checkpoints, we can get high-quality translations of varying complexity, which is useful in settings where data complexity matters. 
For example, guiding teacher model selection for distillation in non-autoregressive machine translation (NAT) can improve the quality of a vanilla NAT model by more than 1 BLEU.

Our contributions are as follows:
\begin{itemize}
    \item we show that during training, NMT undergoes the following three stages:
        \begin{itemize}
        \setlength\itemsep{0em}
            \item[$\circ$] target-side language modeling;
            \item[$\circ$] learning how to use source and approaching word-by-word translation; 
            \item[$\circ$] refining translations, visible by increasingly complex reorderings, but almost invisible to standard metrics (e.g.\ BLEU).
        \end{itemize}
    \item we confirm our finding for different models and modeling paradigms;
    \item we explain how our analysis can be useful in practice and, as an example, show how it can improve a non-autoregressive NMT model.
  
\end{itemize}

\section{Training Stages: The Two Viewpoints}

In this section, we introduce two points of view on the NMT training process. The first one comes from previous work showing distinct stages in NMT training. These stages are formed by looking at a model's internal workings and changes in the way it balances source and target information when forming a prediction. The second point of view is from this work: we take model translations at different training steps and look at some of their aspects mirroring, in a way, core SMT components.

While these two points of view are complete opposites (one sees only the model's innermost workings, the other -- only its output), only taken together they can fully describe the training process. We start from the first, abstract, stages, then show how these inner processes look on the outside and conclude with one of the immediate practical applications of our analysis~(Section~\ref{sect:practical_application}).

\subsection{The Abstract Viewpoint: Relative Token Contributions to NMT Predictions}

\begin{figure}[t!]
    \centering
    {\includegraphics[scale=0.30]{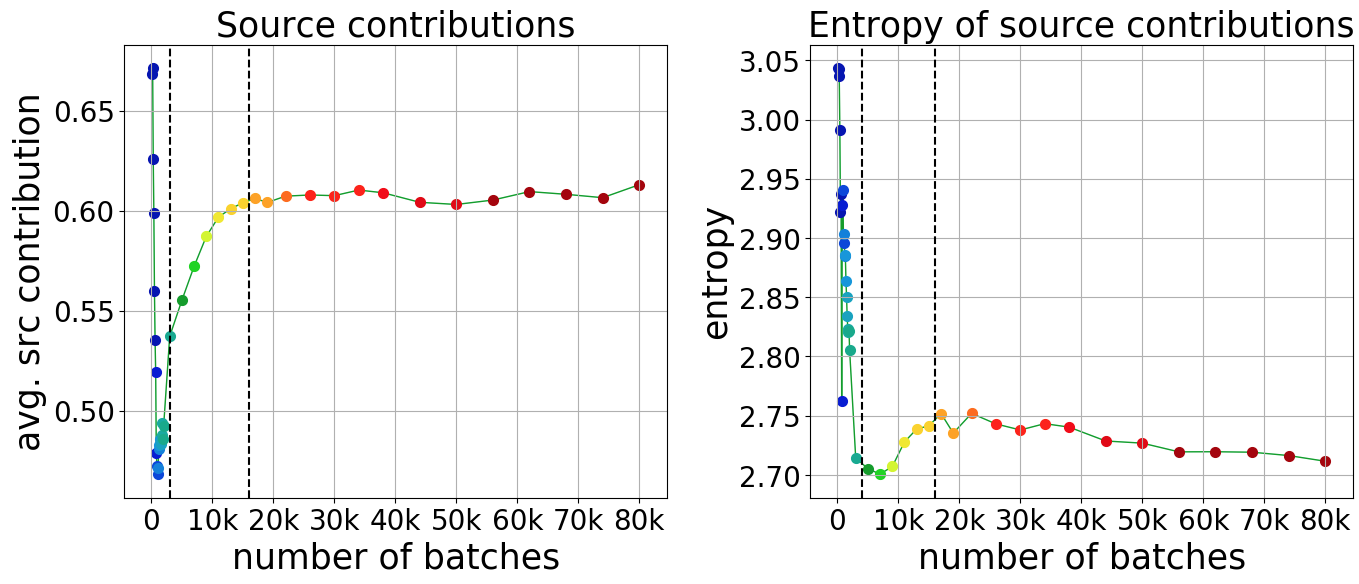}}
  \caption{Contribution of source and entropy of source contributions. En-Ru. Vertical lines separate the stages.}
  \vspace{-2ex}
    \label{fig:lrp_enru}
\end{figure}

The `abstract' stages come from our previous work measuring how NMT balances the two different types of context: the source and prefix of the target sentence~\cite{voita2021analyzing}. We adapt one of the attribution methods, Layerwise Relevance Propagation~\cite{bach2015pixel}, to the Transformer, and show how to evaluate the proportion of each token's influence for a given prediction. Then these relative token influences are used to evaluate the total contribution of the source (by summing up contributions of all source tokens) or to see whether the token contributions are more or less focused (by evaluating the entropy of these contributions).

Among other things, \citet{voita2021analyzing} look at how the total source contribution and the entropy of source contributions change during training. We repeated these experiments for WMT14 En-Ru and En-De.\footnote{Using the released code: \url{https://github.com/lena-voita/the-story-of-heads}.} Figure~\ref{fig:lrp_enru} confirms previous observations: the training process is non-monotonic with several distinct stages, e.g. stages changing direction from decreasing influence of source to increasing.

These results suggest that during training, NMT undergoes stages of qualitatively different changes. For example, a decreasing and then increasing influence of the source likely indicates that the model first learns to rely on the target prefix more (i.e.\ to focus on target-side language modeling) and only after that focuses on the connection to the source (i.e.\ adequacy rather than fluency). While these hypotheses are reasonable, to confirm them we have to look not only at how model predictions are formed but also at the predictions themselves.

\subsection{The Practical Viewpoint: Model Translations}

In this viewpoint, we are interested in changes in model output, i.e. translations. We measure:
\begin{itemize}
\setlength\itemsep{-0.2em}
    \item[$\circ$] target-side language modeling scores;
    \item[$\circ$] translation quality; 
    \item[$\circ$] monotonicity of alignments. 
\end{itemize}
Note that these characteristics are related to three core components of the traditional SMT models: target-side language model, translation model, and reordering model. Although we are mainly interested in NMT models and, except for the language modeling scores, do not measure the quality of the corresponding SMT components directly, this relation to SMT is important. While machine translation is now mostly neural, it is still not clear how (e.g., in which order) those competences which used to be modelled with distinct components are now learned jointly within a single neural network.

\section{Experimental Setting}
\label{sect:experimental_setting}

\subsection{Models, Data and Preprocessing}

\paragraph{Models.} We consider three models:
\begin{itemize}
\setlength\itemsep{-0.2em}
    \item[$\circ$] Transformer encoder-decoder;
    \item[$\circ$] LSTM encoder-decoder;
    \item[$\circ$] Transformer decoder (LM-style NMT).
\end{itemize}
For the first model, we follow the setup of the Transformer base~\cite{attention-is-all-you-need}. LSTM encoder-decoder is a single-layer GNMT~\cite{wu2016googles}. The last model is the Transformer decoder trained as a left-to-right language model. In training, the model receives concatenated source and target sentences separated by a token-delimiter; in inference, it receives only the source sentence and the delimiter and is asked to continue generation.

\paragraph{Datasets.} We use the WMT news translation shared task for English-German and English-Russian: for En-De, WMT 2014 with 5.8m sentence pairs, for En-Ru~-- 2.5m sentence pairs (parallel training data excluding UN and Paracrawl). Since our observations are similar for both languages, in the main text we show figures for one of them and in the appendix~-- for the other.

\paragraph{Preprocessing.} The data is lowercased and encoded using BPE~\cite{sennrich-bpe}. We use separate source and target vocabularies of about 32k tokens for encoder-decoder models, and a joint vocabulary of about 50k tokens for LM-style models.
For each experiment, we randomly choose 2/3 of the dataset for training and use the remaining 1/3 as a held-out set for analysis~(see Section~\ref{sect:introduce_reordering_score}).

More details on hyperparameters, preprocessing, and training can be found in the appendix.

\subsection{Target-Side LM Scores}

For each of the models, we train 2-, 3-, 4- and 5-gram
KenLM~\cite{heafield-2011-kenlm}\footnote{\url{https://github.com/kpu/kenlm}} language models on target sides of the corresponding training data (segmented with BPE). We report KenLM scores for the translations of the development sets.

\begin{figure*}[t!]
    \centering
    \subfloat[]
    {\includegraphics[scale=0.32]{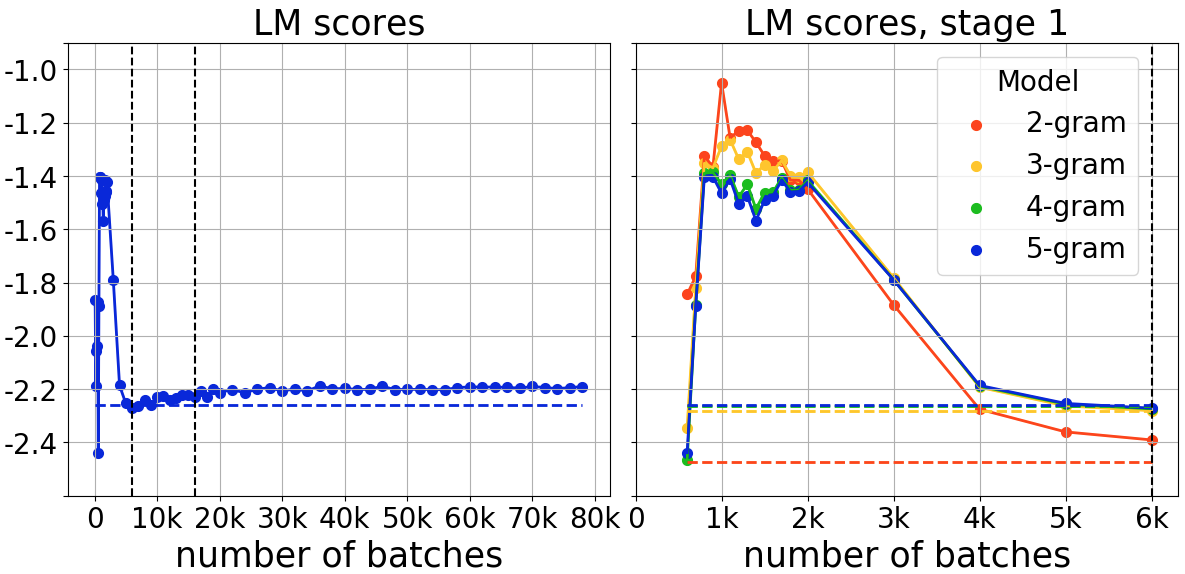}}
    \quad\quad
    \subfloat[]
    {\includegraphics[scale=0.32]{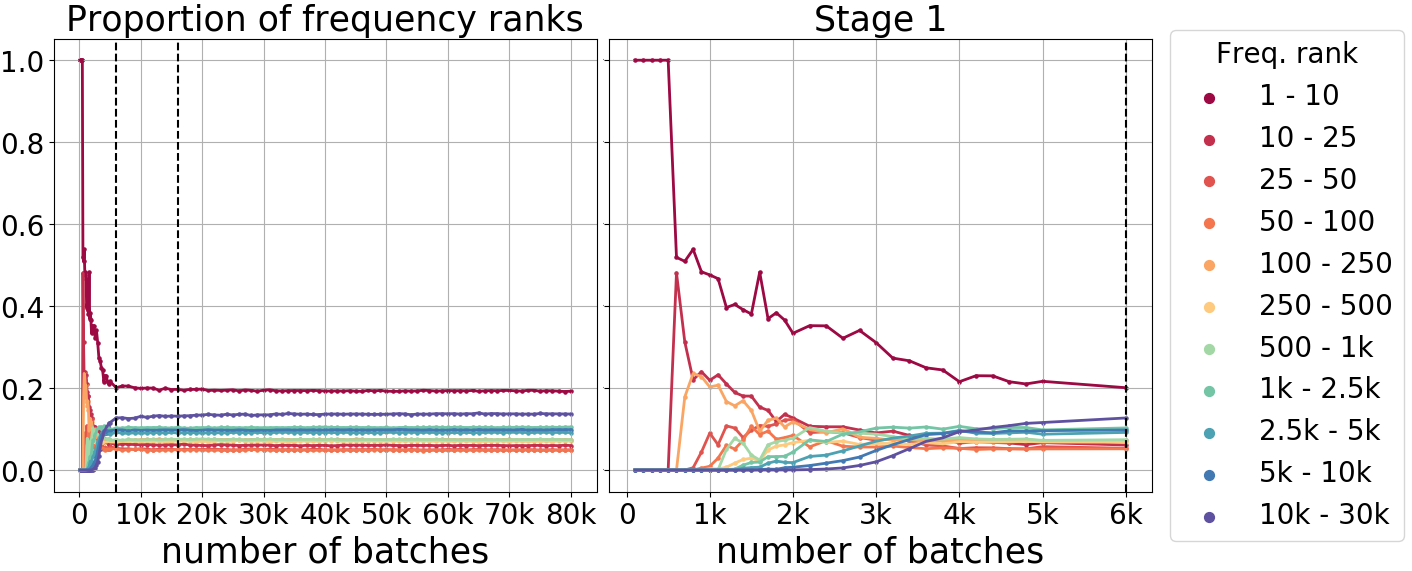}}
    
  \caption{(a) KenLM scores (horizontal dashed lines are the scores for the references); (b) proportion of tokens of different frequency ranks in model translations. En-Ru.}
    \label{fig:lm_and_freq_ranks}
\end{figure*}

\begin{figure*}[t!]
{\includegraphics[scale=0.24]{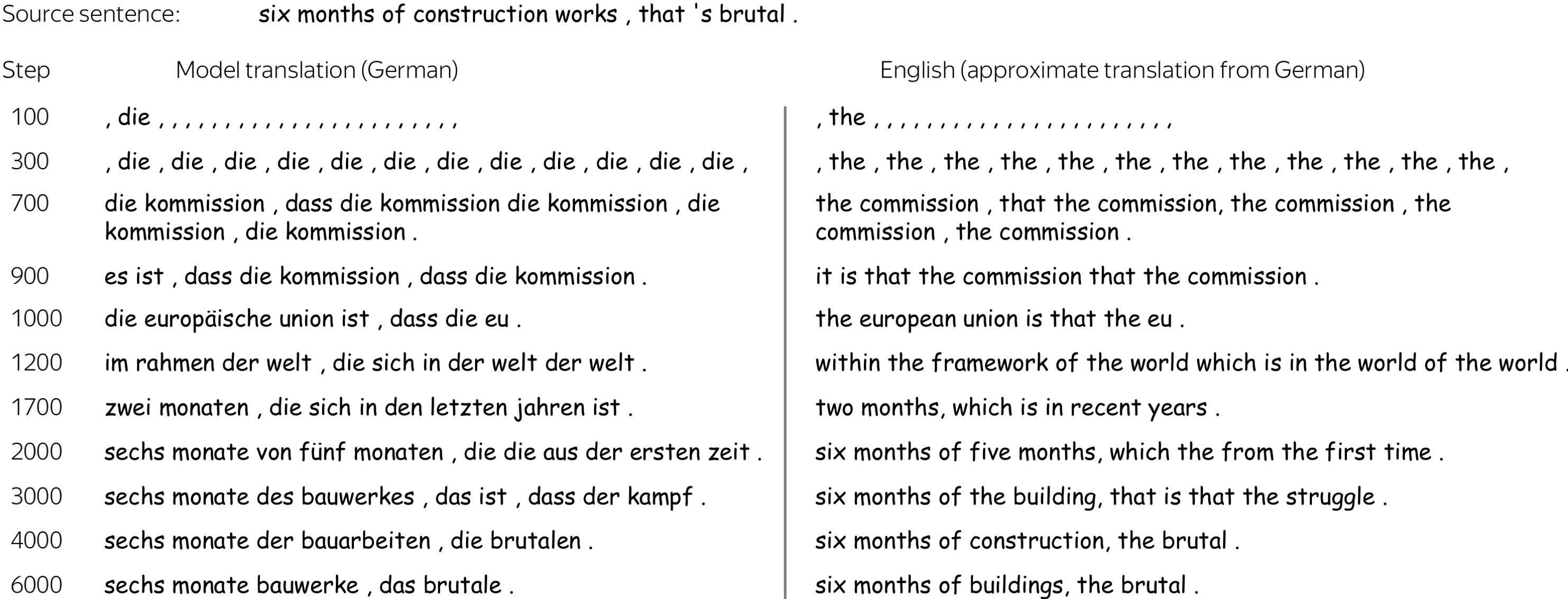}}
\caption{Translations at different steps during training. En-De.}
\label{fig:lm_stage_examples_de}
\end{figure*}

\subsection{Monotonicity of Alignments}
\label{sect:introduce_reordering_score}

To measure how the relative ordering of words in the source and its translation changes during training, we use two different scores used in previous work~\cite{burlot-yvon-2018-using,Zhou2020Understanding}. We evaluate the scores for two permutations of the source: the trivial monotonic alignment and the alignment inferred for the generated translation.

\paragraph{Fuzzy Reordering Score}~\cite{talbot-etal-2011-lightweight} counts the number of chunks of contiguously aligned words and, intuitively, it is based on the number of times a reader would need to jump in order to read one reordering in the order proposed by the other. The score is between 0 and 1, where a larger score indicates more monotonic alignments.

\paragraph{Kendall tau distance}~\cite{kendal-tau} is also called \textit{bubble-sort distance} since it is equivalent to the number of swaps that the bubble sort algorithm would take to place one list in the same order as the other list. We evaluate the normalized distance: it is between 0 and 1, where 0 indicates the monotonic alignment.

The main difference between the scores is that the first one takes into account only the number of jumps, while the second also considers their distance. For a formal description of the scores and their differences, see the appendix.

\paragraph{Our setting.} For each of the considered  model checkpoints, we obtain datasets where the sources come from the held-out 1/3 of the original dataset, and targets are their translations. For these datasets, we infer alignments using \texttt{fast\_align}~\cite{dyer-etal-2013-simple}\footnote{\url{https://github.com/clab/fast_align}}.

\section{Transformer Training Stages}
\label{sect:transformer_training_stages}

In this section, we discuss the standard encoder-decoder Transformer. In the next section, we mention differences with several other models.

We first analyze the results for each of the three competences and then characterize the stages based on these practical observations. In all figures, we show the abstract stages with vertical lines to link the results to the changes in token contributions.



\subsection{Target-Side Language Modeling}

Figure~\ref{fig:lm_and_freq_ranks}a shows changes in the language modeling scores. We see that most of the change happens in the very beginning: the scores go up and peak much higher than that of the references. This means that the model generates sentences with very frequent n-grams rather than diverse texts similar to references. Indeed, Figure~\ref{fig:lm_and_freq_ranks}a (right) shows that for a part of the training (from 1k to 2k iterations), the scores for simpler models (e.g., 2-gram) are higher than for the more complicated ones (e.g., 5-gram). This means that generated translations tend to consist of frequent words and bigrams, but larger subsequences are not necessarily fluent.

To illustrate this, we show how translations of one of the sentences evolve at the beginning of training~(Figure~\ref{fig:lm_stage_examples_de}). As expected, first the translations evolve from repetitions of frequent tokens to frequent bigrams and trigrams, and finally to longer frequent phrases. To make this more clear, we also show the proportion of tokens of different frequency ranks in generated translations~(Figure~\ref{fig:lm_and_freq_ranks}b). First (iterations 0-500), all generated tokens are from the top-10 most frequent tokens, then only from the top-50, and only later less frequent tokens are starting to appear. From Figure~\ref{fig:lm_stage_examples_de} we see that this happens when the source comes into play: tokens related to the source become weaved into translations. Overall, this evolution from using short target-side contexts to longer ones and, subsequently, to using the source relates to works in computer vision discussing `shortcut features'~\cite{DBLP:journals/corr/abs-2004-07780}, as well as differences in the progression of extracting `easy' and `difficult' features during training~\cite{NEURIPS2020_71e9c662}.

Note also that model translations converge to higher LM scores than references (Figure~\ref{fig:lm_and_freq_ranks}a). This is expected: compared to references, beam search translations are simpler in various aspects, e.g. they are simpler syntactically, contain fewer rare tokens and less reordering~\cite{burlot-yvon-2018-using,pmlr-v80-ott18a,Zhou2020Understanding}, and lead to more confident token contributions inside the model~\cite{voita2021analyzing}. For language models more generally, beam search texts are also less surprising than human ones~\cite{Holtzman2020The}.

To summarize, the beginning of training is mostly devoted to target-side language modeling: we see huge changes in the LM scores (Figure~\ref{fig:lm_and_freq_ranks}a), and the model hallucinates frequent n-grams (Figure~\ref{fig:lm_stage_examples_de}). This agrees with the abstract stages shown in Figure~\ref{fig:lrp_enru}: in the first stage, the total contribution of the source substantially decreases. This means that in the trade-off between information coming from the source and the target prefix, the model gives more and more priority to the prefix.


\subsection{Translation Quality}
\label{sect:main_translation_quality}

\begin{figure}[t!]
    \centering
    \subfloat[]
    {\includegraphics[scale=0.19]{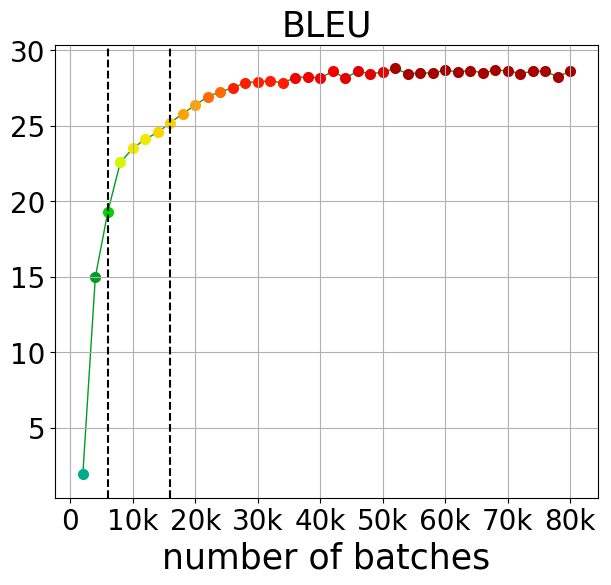}}
    \ 
    \subfloat[]
    {\includegraphics[scale=0.28]{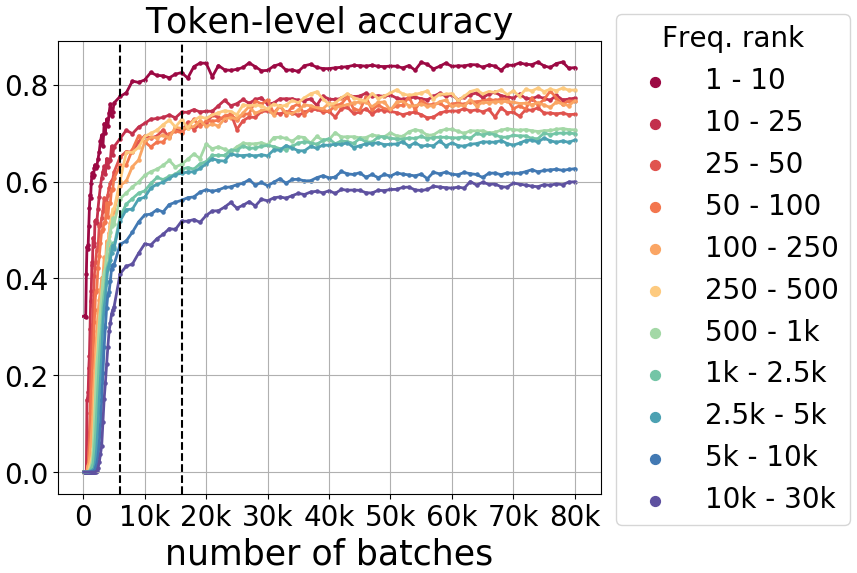}}
    \vspace{-1ex}
  \caption{(a) BLEU score; (b) token-level accuracy (the proportion of cases where the correct next token is the most probable choice). WMT En-Ru.}
  \vspace{-2ex}
    \label{fig:enru_bleu_acc}
\end{figure}

Figure~\ref{fig:enru_bleu_acc}a shows the BLEU score on the development set during training. For a more fine-grained analysis, we also plot token-level predictive accuracy separately for target token frequency groups~(Figure~\ref{fig:enru_bleu_acc}b). We see that both the BLEU score and accuracy become large very fast, e.g. after the first 20k iterations (25$\%$ of the training process), the scores are already good. What is interesting, is that the accuracy for frequent tokens reaches the maximum value (the score of the converged model) very quickly. This agrees with our previous observations in Figures~\ref{fig:lm_stage_examples_de} and~\ref{fig:lm_and_freq_ranks}b: at the beginning of training, the model generates frequent tokens more readily than the rare ones. Figure~\ref{fig:enru_bleu_acc}b further confirms this: the accuracy for the rare tokens improves slower than for the rest of them.

What is not clear, is what happens during the last half of the training (iterations from 40k to 80k): BLEU score improves only by 0.4, accuracy does not seem to change noticeably even for rare tokens, the proportion of generated tokens of different frequency ranks converges even earlier (Figure~\ref{fig:lm_and_freq_ranks}b), and patterns in token contributions also do not change much~(Figure~\ref{fig:lrp_enru}). This is what we are about to find out in the next section.

\begin{figure}[t!]
    \centering
    \subfloat[]
    {\ \ \includegraphics[scale=0.20]{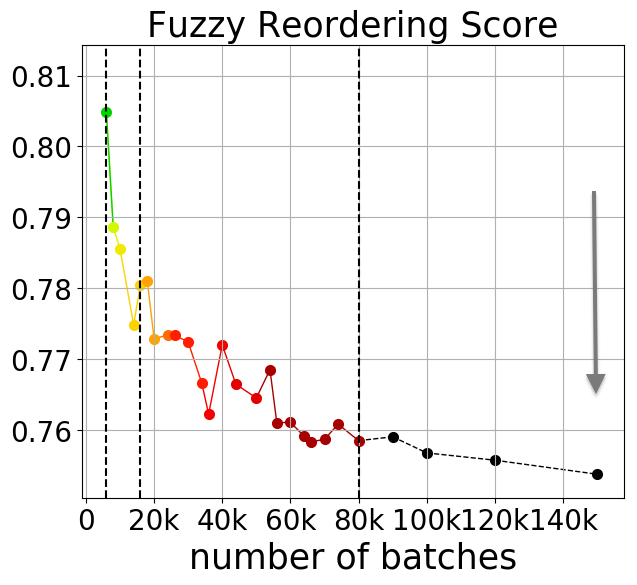}}
    \quad
    \subfloat[]
    {\includegraphics[scale=0.20]{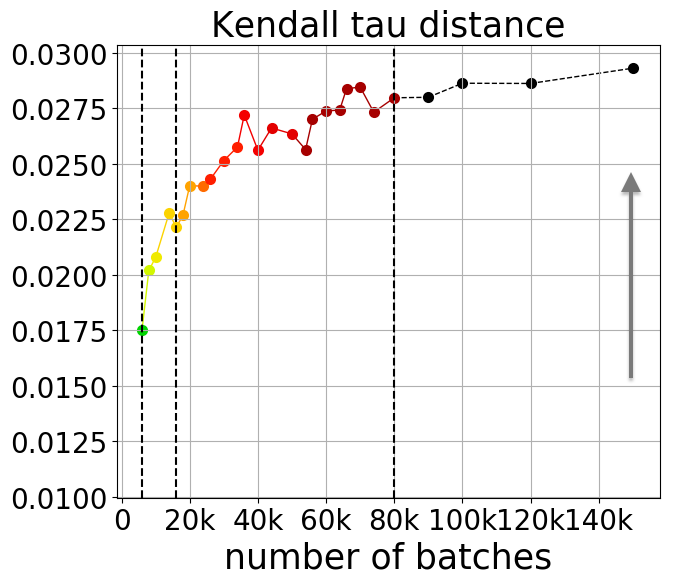}}
    \vspace{-1ex}
  \caption{(a) fuzzy reordering score (for references: 0.6), (b) Kendall tau distance (for references: 0.06); WMT En-Ru. The arrows point in the direction of less monotonic alignments (more complicated reorderings).}
  \vspace{-2ex}
    \label{fig:enru_frs}
\end{figure}

\begin{figure*}[t!]
\subfloat[En-De]
{\includegraphics[scale=0.24]{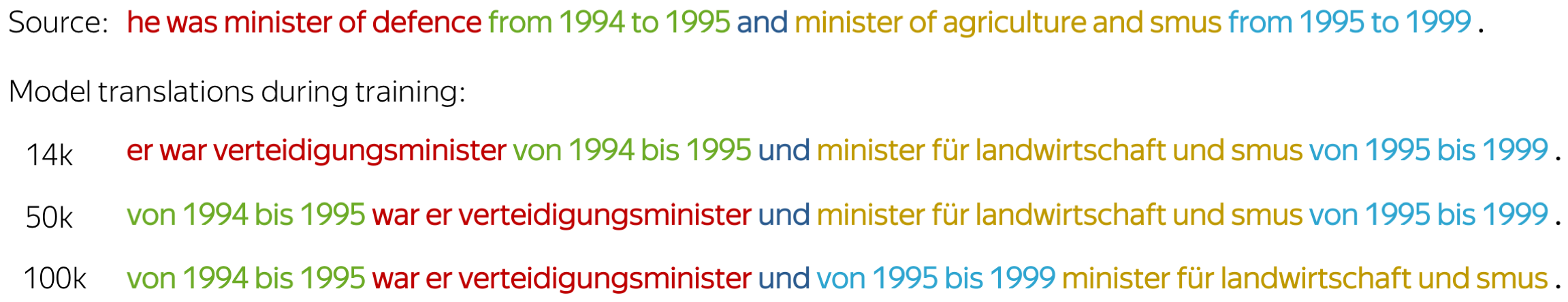}}
\\
\subfloat[En-Ru]
{\includegraphics[scale=0.24]{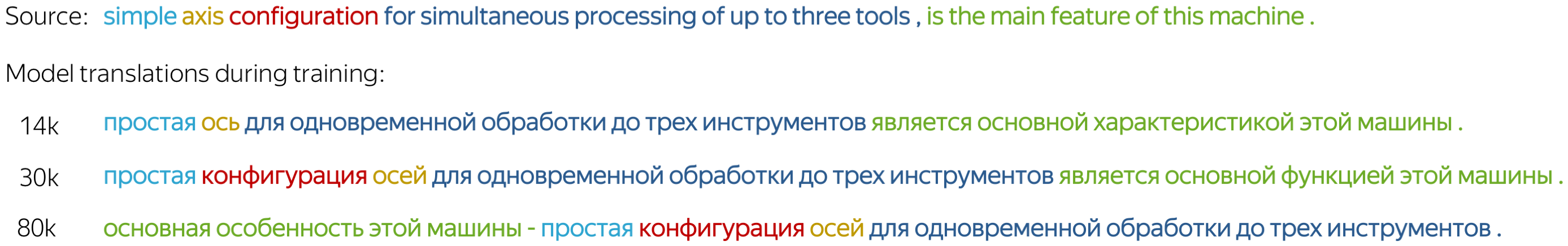}}
\vspace{-1ex}
\caption{Translations at different training steps. Same-colored chunks are approximately aligned to each other.}
\vspace{-2ex}
\label{fig:reordering_stage_examples}
\end{figure*}

\subsection{Monotonicity of Alignments}
\label{sect:monotonicity}

While it is known that, compared to references, beam search translations have more monotonic alignments~\cite{burlot-yvon-2018-using,Zhou2020Understanding}, it is not clear how monotonicity of alignments changes during model training. We show changes in the two reordering scores in Figure~\ref{fig:enru_frs}.\footnote{Note that we evaluate the scores starting not from the very beginning of training but after at least 6k updates. This is because evaluating monotonicity of alignments makes sense only when translations are reasonable.}

We can say that during the second half of the training,
the model is slowly refining translations, and, among the three competences we look at, the most visible changes are due to more complicated (i.e. less monotonic) reorderings. For example, as we already mentioned above, during this part of the training none of the scores we looked at so far changes much, whereas changes in both reordering scores are very substantial. The change in the fuzzy reordering score is only twice smaller than during the preceding stage. Moreover, the alignments keep changing and become less monotonic even after both BLEU and token-level accuracy (i.e. the metric that matches the model's training objective) converged, i.e. iterations after 80k (Figure~\ref{fig:enru_frs}). 

Overall, we interpret this refinement stage as the model slowly learning to reduce interference from the source text (typical for human translation~\cite{Volansky2015OnTF} and exacerbated even more in NMT~\cite{toral-2019-post}): it learns to apply complex reorderings to more closely follow typical word order in the target language. This means that while language modeling improves more prominently during the first training stage, there is a long tail of less frequent and more nuanced patterns that the model learns later.
Another example of such nuanced changes in translation not detected with standard metrics is context-aware NMT. 
Previous work has criticized using BLEU as a stopping criterion, showing that even when a model has converged in terms of BLEU, it continues to improve in terms of agreement with context~\cite{voita-etal-2019-good}.

To illustrate changes during this last stage, we show two examples in Figure~\ref{fig:reordering_stage_examples}. On average, the translations at the beginning of the last stage tend to have the same word order as the corresponding source sentences: the  alignments are highly monotonic. Formally, the similarity to the word-by-word translation is seen from the very low Kendall tau distance after 6k-14k training iterations (Figure~\ref{fig:enru_frs}b): this means that a very small number of permutations is needed to transform the trivial monotonic translation into the one produced by the model. Interestingly, at this point, some undertranslation errors can be explained via failures to perform a complex reordering.
 In the example in Figure~\ref{fig:reordering_stage_examples}b, the phrase `axis configuration' cannot be translated into Russian preserving the same word order, which makes the model to omit the translation of `configuration'.

\subsection{Characterizing Training Stages}
\label{sect:characterizing_training_stages}

To summarize, the NMT training process can be described as undergoing the following three stages:
\begin{itemize}
\setlength\itemsep{-0.2em}
    \item[$\circ$] target-side language modeling;
    \item[$\circ$] learning how to use source and coming close to a word-by-word translation; 
    \item[$\circ$] refining translations, visible by an increase in complexity of the reorderings and almost invisible by standard evaluation (e.g. BLEU).
\end{itemize}
While the borders of these practical stages are not as strictly defined as the abstract ones with the changes of monotonicity in contribution graphs~(Figure~\ref{fig:lrp_enru}), these two points of view on the training process mirror each other very well. From the abstract point of view with token contributions, the model first starts to form its predictions based more on the prefix and ignores the source, then source influence increases quickly, then very little is going on~(Figure~\ref{fig:lrp_enru}). From the practical point of view with model translations, the model first hallucinates frequent tokens, then phrases, then sentences (mirrors source contributions going down), then quickly improves translation quality (mirrors source contribution going up), then little is going on according to the standard scores, but alignments become noticeably less monotonic. As we see, both points of view show the same kinds of processes from different perspective: from the inside and the outside of the model.

\section{Other NMT Models}

In this section, we compare different architectures within the same encoder-decoder framework (Transformer vs LSTM), and different frameworks with the Transformer architecture (encoder-decoder vs decoder-only).
Overall, we find that all models follow the behavior described in Section~\ref{sect:characterizing_training_stages}; here we discuss some of their differences.

\begin{figure}[t!]
    \centering
    \subfloat[]
    {\includegraphics[scale=0.23]{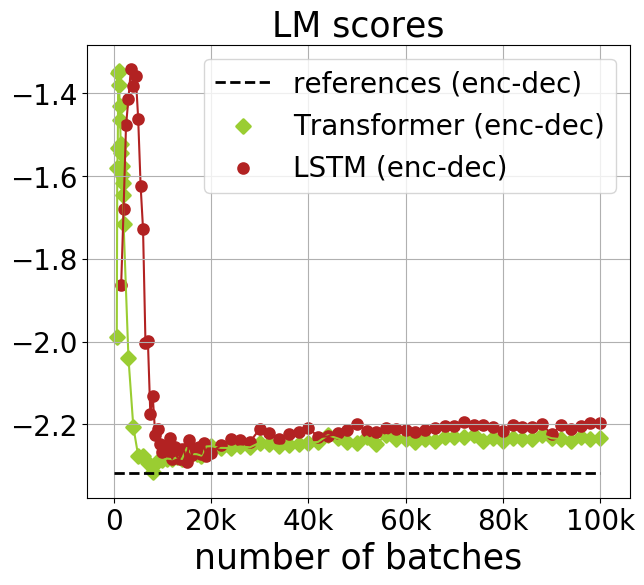}}
    \ \ 
    \subfloat[]
    {\includegraphics[scale=0.23]{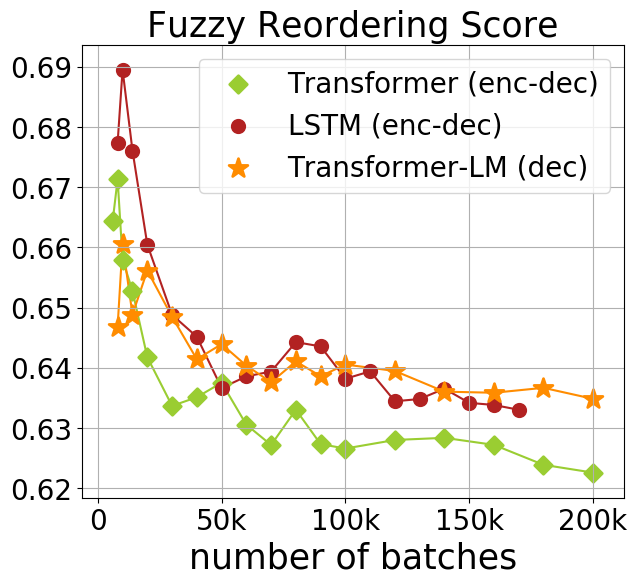}}
    \vspace{-1ex}
  \caption{(a) target-side LM scores (5-gram), (b) fuzzy reordering score (for references: 0.5); WMT En-De.}
  \vspace{-2ex}
    \label{fig:other_models_ende}
\end{figure}

\paragraph{Transformer vs LSTM.} As might be expected from the low BLEU scores (Table~\ref{tab:bleu_scores}), LSTM translations are simpler than the Transformer ones. We see that they are less surprising according to the target-side language modeling scores (Figure~\ref{fig:other_models_ende}a\footnote{Note that in Figure~\ref{fig:other_models_ende}a, only the scores of the encoder-decoder models can be compared because of differences in model vocabulary (see Section~\ref{sect:experimental_setting}). In the appendix, we show scores for all three models.}) and have more monotonic alignments (Figure~\ref{fig:other_models_ende}b). Regarding the latter, it is not clear whether this is because of the lower model capacity or because LSTM has an inductive bias towards more monotonic alignments; we leave this to future work.

\paragraph{Encoder-decoder vs decoder-only.} Table~\ref{tab:bleu_scores} shows that decoder-only (LM-style) NMT is not much worse than the standard encoder-decoder model, especially in the higher-resource setting (e.g., En-De). However, the decoder-only model has much simpler reordering patterns compared to the standard Transformer: its reordering scores are very close to the much weaker LSTM model (Figure~\ref{fig:other_models_ende}b). One possible explanation is that the bidirectional nature of Transformer's encoder facilitates learning more complicated reorderings.

\begin{table}[t!]
\centering
\begin{tabular}{lcc}
\toprule
 \bf model & \bf En-Ru & \bf En-De \\ 
 \cmidrule{1-3}
Transformer (enc-dec)   &  35{.}93 & 28{.}18 \\
LSTM (enc-dec) & 30{.}14 & 24{.}03\\
Transformer-LM (dec) & 34{.}16 & 26{.}76 \\
\bottomrule
\end{tabular}\textbf{}
\caption{BLEU scores: \texttt{newstest2014} for En-Ru and \texttt{newstest2017} for En-De.}
\vspace{-2ex}
\label{tab:bleu_scores}
\end{table}

\section{Practical Implications}
\label{sect:practical_application}

We showed that during a large part of the training, the translation quality (e.g., BLEU) changes little, but the alignments become less monotonic. Intuitively, the translations become more complicated while their quality remains roughly the same.

One way to directly apply our analysis is to consider tasks and settings where data properties such as regularity and/or simplicity are important. For example, in neural machine translation, 
higher monotonicity of artificial sources was hypothesized to be a facilitating factor for back-translation~\cite{burlot-yvon-2018-using}; additionally, complexity of the distilled data is crucial for sequence-level distillation in non-autoregressive machine translation~\cite{Zhou2020Understanding}. Such examples are not limited to machine translation: in emergent languages, 
languages with higher `regularity' bring learning speed advantages for communicating neural agents~\cite{Ren2020Compositional}.

In this section, we consider non-autoregressive NMT, and leave the rest to future work.

\subsection{Non-Autoregressive Machine Translation}

Non-autoregressive neural machine translation~(NAT) \citep{gu2018nonautoregressive} is different from the traditional NMT in the way it generates target sequences: instead of the standard approach where target tokens are predicted step-by-step by conditioning on the previous ones, NAT models predict the whole sequence simultaneously. This is possible only with an underlying assumption that the output tokens are independent from each other, which is unrealistic for natural language.

Fortunately, while this independence assumption is unrealistic for real references, it might be more plausible for simpler sequences, e.g. artificially generated translations. That is why targets for NAT models are usually not references but beam search translations of the standard autoregressive NMT (which, as we already mentioned above, are simpler than references in many aspects). 
This is called \textit{sequence-level knowledge distillation}~\cite{kim-rush-2016-sequence}, and it is currently one of the de-facto standard parts of the NAT training pipelines~(\citet{gu2018nonautoregressive,lee-etal-2018-deterministic,ghazvininejad-etal-2019-mask} to name a few).

Recently~\citet{Zhou2020Understanding} showed that the quality of a NAT model strongly depends on the complexity of the distilled data, and changing this complexity can improve the model. Since distilled data consists of translations from a standard autoregressive teacher, our analysis gives a very simple way of modifying the complexity of this data. While usually a teacher is a fully converged model, we propose to use as teachers intermediate checkpoints during training. Since during a large part of training, NMT quality (e.g., BLEU) changes little, but the alignments become less monotonic, earlier checkpoints can produce simpler and more monotonic translations. We hypothesize that these translations are more suitable as targets for NAT models, and we confirm this with the experiments.

\subsection{Setting}

Following previous work~\cite{Zhou2020Understanding}, we train the same NAT model on their preprocessed dataset\footnote{We used the code and the data from \url{https://github.com/pytorch/fairseq/tree/master/examples/nonautoregressive_translation}.} and vary only distilled targets.

\paragraph{Model.} The model is the re-implemented by~\citet{Zhou2020Understanding} version of the vanilla NAT by~\citet{gu2018nonautoregressive}. For more details, see appendix. 

\paragraph{Dataset.} The dataset is WMT14 English-German (En-De) with newstest2013 as the validation set and newstest2014 as the test set, and BPE vocabulary of 37,000. We use the preprocessed dataset and the vocabularies released by~\citet{Zhou2020Understanding}.

\paragraph{Distilled targets.} The teacher is the standard Transformer-base from~\texttt{fairseq}~\cite{fairseq}. For the baseline distilled dataset, we use the fully converged model (in this case, the model after 200k updates). For other datasets, we use earlier checkpoints. 

\paragraph{Evaluation.} We average the last 10 checkpoints.

\subsection{Experiments}

\begin{figure}[t!]
    \centering
    \subfloat[]
    {\includegraphics[scale=0.25]{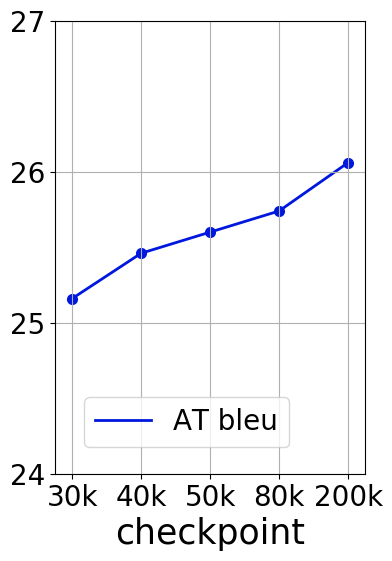}}
    \subfloat[]
    {\includegraphics[scale=0.25]{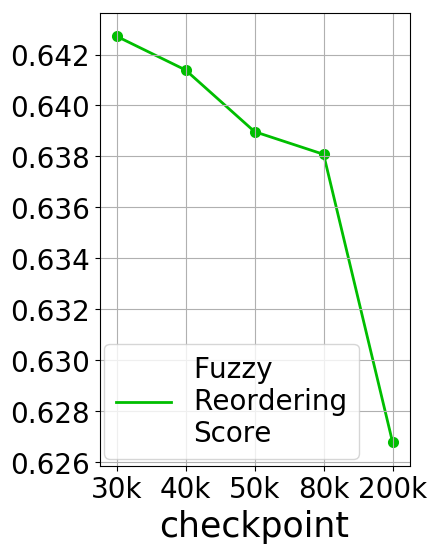}}
    \subfloat[]
    {\includegraphics[scale=0.11]{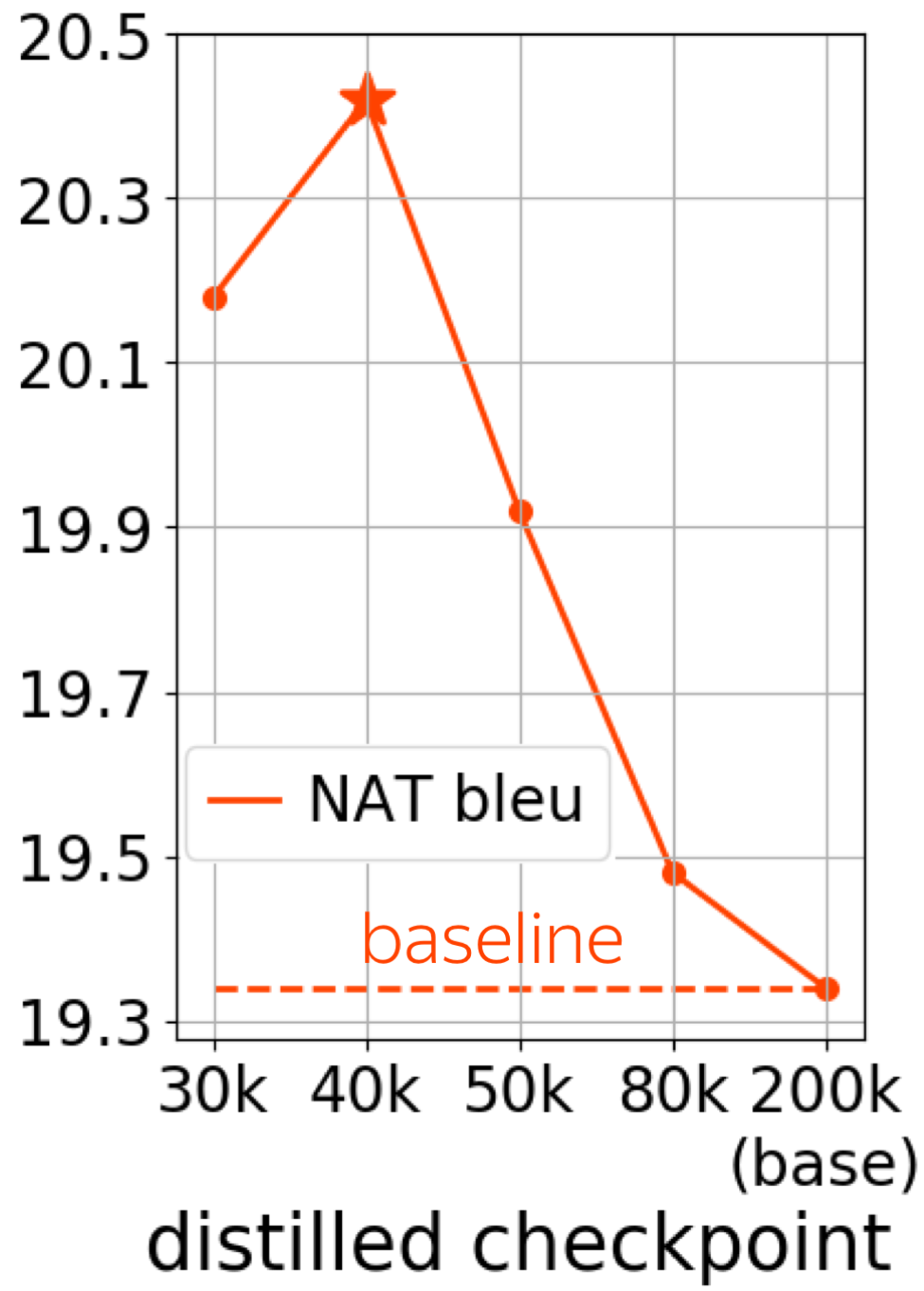}}
    \vspace{-1ex}
  \caption{(a) BLEU score of the AT Transformer-base (teacher for distillation); (b) fuzzy reordering score for the distilled training data obtained from checkpoints of the AT teacher; (c) BLEU scores for the vanilla NAT model trained on different distilled data.}
  \vspace{-2ex}
    \label{fig:nat_results}
\end{figure}

Figure~\ref{fig:nat_results}c shows the BLEU scores for NAT models trained with distilled data obtained from different teacher's checkpoints; the baseline is the fully converged model (200k iterations). We see that by taking an earlier checkpoint, after 40k iterations, we improve NAT quality by 1{.}1 BLEU. For this checkpoint, the teacher's BLEU score is not much lower than that of the final model~(Figure~\ref{fig:nat_results}a), but the reorderings are much simpler~(a higher fuzzy reordering score in Figure~\ref{fig:nat_results}b). 

To vary the complexity of the distilled data, \citet{Zhou2020Understanding} proposed to apply either Born-Again networks~(BANs)~\cite{pmlr-v80-furlanello18a} or  mixture-of-experts~(MoE)~\cite{pmlr-v97-shen19c}. Unfortunately, MoE is rather complicated and requires careful hyperparameter tuning~\cite{pmlr-v97-shen19c}, and BANs are time- and resource-consuming. They involve training the AT model till convergence and then translating the training data to get a distilled dataset; this happens in several iterations (e.g., 5-7) using for training the latest generated dataset. Compared to these methods, our approach is extremely simple and does not require a lot of computational resources (e.g., instead of fully training the AT teacher several times as in BANs, our approach requires only to partially train one AT teacher). 

Note that in this work, 
we provide these experiments mainly to illustrate how our analysis can be useful in the settings where data complexity matters and, therefore, limit ourselves to only using different teacher checkpoints. Future work, however, can  investigate possible combinations with other approaches. For example, to further improve quality, our method can be combined with the Born-Again networks while still requiring fewer resources due to only partial training of the teachers.

\section{Additional Related Work}
\label{sect:related_work}

Other work connecting neural and traditional approaches include modeling modifications, such as modeling coverage and/or fertility~\cite{tu-etal-2016-modeling,mi-etal-2016-coverage,cohn-etal-2016-incorporating,feng-etal-2016-improving} and several other modifications~\cite{zhang-etal-2017-improving,stahlberg-etal-2017-neural,huang2018towards}, analysis of the relation between attention and word alignments~\cite{ghader-monz-2017-attention}, and word alignment induction from NMT models~\cite{li-etal-2019-word,garg-etal-2019-jointly,Song2020TowardsBW,zenkel-etal-2020-end,chen-etal-2020-accurate}.

Previous analysis of NMT learning dynamics
include analyzing how the trainable parameters affect an NMT model~\cite{zhu2020understanding} and looking at the speed of learning specific discourse phenomena in context-aware NMT~\cite{voita-etal-2019-good,voita-etal-2019-context}.

\section{Conclusions}

We analyze how NMT acquires different competencies during training and look at the competencies related to three core SMT components. We find that NMT first focuses on learning target-side language modeling, then improves translation quality approaching word-by-word translation, and finally learns more complicated reordering patterns. We show that such an understanding of the training process can be useful in settings where data complexity matters and illustrate this for non-autoregressive MT; other tasks can be considered in future work. Additionally, our results can contribute to the discussion of (i) `easy' and `difficult' task-relevant features, including `shortcut features', and (ii) the limitations of the BLEU score.

\section*{Acknowledgments} We would like to thank the anonymous reviewers for their comments. Lena is supported by the Facebook PhD Fellowship. Rico Sennrich acknowledges support of the Swiss National Science Foundation (MUTAMUR; no. 176727). Ivan Titov acknowledges support of the European Research Council (ERC StG BroadSem 678254), Dutch National Science Foundation (VIDI 639.022.518) and EU Horizon 2020 (GoURMET, no. 825299).

\bibliography{emnlp2021}
\bibliographystyle{acl_natbib}

\newpage
\phantom{0}
\newpage

\appendix

\input{appendix.tex}

\end{document}

%% file: appendix.tex
\section{Experimental Setting}

\subsection{Data preprocessing}

Translation pairs were batched together by approximate sequence length. Each training batch contained a set of translation pairs containing approximately 32000 source tokens.\footnote{This can be reached by using several of GPUs or by accumulating the gradients for several batches and then making an update.}

\subsection{Model parameters}

\paragraph{Transformer (encoder-decoder).} We follow the setup of Transformer base model~\cite{attention-is-all-you-need}. More precisely, the number of layers in the encoder and in the decoder is $N=6$. We employ $h = 8$ parallel attention layers, or heads. The dimensionality of input and output is $d_{model} = 512$, and the inner-layer of a feed-forward networks has dimensionality $d_{ff}=2048$. We use regularization as described in~\cite{attention-is-all-you-need}.

\paragraph{Transformer (decoder).} The difference from the previous model is that the decoder has 12 layers.

\paragraph{LSTM (encoder-decoder)} is a single-layer GNMT~\cite{wu2016googles} with the input and output dimensionality of 512 and hidden sizes of 1024.

\subsection{Optimizer}

The optimizer we use is the same as in~\cite{attention-is-all-you-need}.
We use the Adam optimizer~\cite{adam-optimizer} with $\beta_1 = 0{.}9$, $\beta_2 = 0{.}98$ and $\varepsilon = 10^{-9}$. We vary the learning rate over the course of training, according to the formula:
$$
l_{rate}=scale\cdot \min(step\_num^{-0.5},$$ $$step\_num\cdot warmup\_steps^{-1.5}) 
$$
We use $warmup\_steps = 16000$, $scale=4$.

\section{Monotonicity of Alignments}

To measure how the relative ordering of words in the source and target sentences changes during training, we use two different scores: fuzzy reordering score~\cite{talbot-etal-2011-lightweight} and Kendall tau distance. We evaluate both scores for two permutations of the source sentence $\sigma_1$ and $\sigma_2$, where $\sigma_1$is the trivial monotonic alignment and $\sigma_2$ -- the alignment inferred for the generated translation.

\paragraph{Fuzzy Reordering Score} aligns each word in $\sigma_1$ to an instance of itself in $\sigma_2$ taking the first unmatched instance of the word if there is more than one. If $C$ is the number of chunks of contiguously aligned words and $M$ is the number of words in the source sentence, then the fuzzy reordering score is computed as 
\begin{equation}
FRS(\sigma_1, \sigma_2) = 1 - \frac{C-1}{M-1}.
\label{eq:fuzzy_reorderin_score}
\end{equation}
This metric assigns a score between 0 and 1, where 1 indicates that the two reorderings are identical. Intuitively, $C$ is the number of times a reader would need to jump in order to read the reordering $\sigma_1$ in the order proposed by $\sigma_2$. A larger fuzzy reordering score indicates more monotonic alignments.

\paragraph{Kendall tau distance} counts the number of pairwise disagreements between two ranking lists. The larger the distance, the more dissimilar the two lists are. Kendall tau distance is also called \textit{bubble-sort distance} since it is equivalent to the number of swaps that the bubble sort algorithm would take to place one list in the same order as the other list. We evaluate the normalized distance, i.e. for a list of length $n$ it is normalized by $\frac{n(n-1)}{2}$. The normalized score is between 0 and 1, where 0 indicates that the two reorderings are identical.

\paragraph{Differences between the scores.} While the first score counts only the number of chunks of contiguously aligned words, the second one takes into account only how distant the changes are. For example, let us consider two reorderings: $(2, 1, 4, 3, 6, 5)$
and $(4, 5, 6, 1, 2, 3)$. While for the fuzzy reordering score the least monotonic reordering is the first (more jumps for a reader), for the Kendall tau score -- the second (requires more permutations to reorder). As we will see in Section~\ref{sect:monotonicity}, results for the two scores are similar.

\paragraph{Our setting.} We take sentences of at least 2 words for the fuzzy reordering score and at least 10 tokens for the Kendall tau distance.

\section{Transformer Training Stages}

\begin{figure}[t!]
    \centering
    {\includegraphics[scale=0.31]{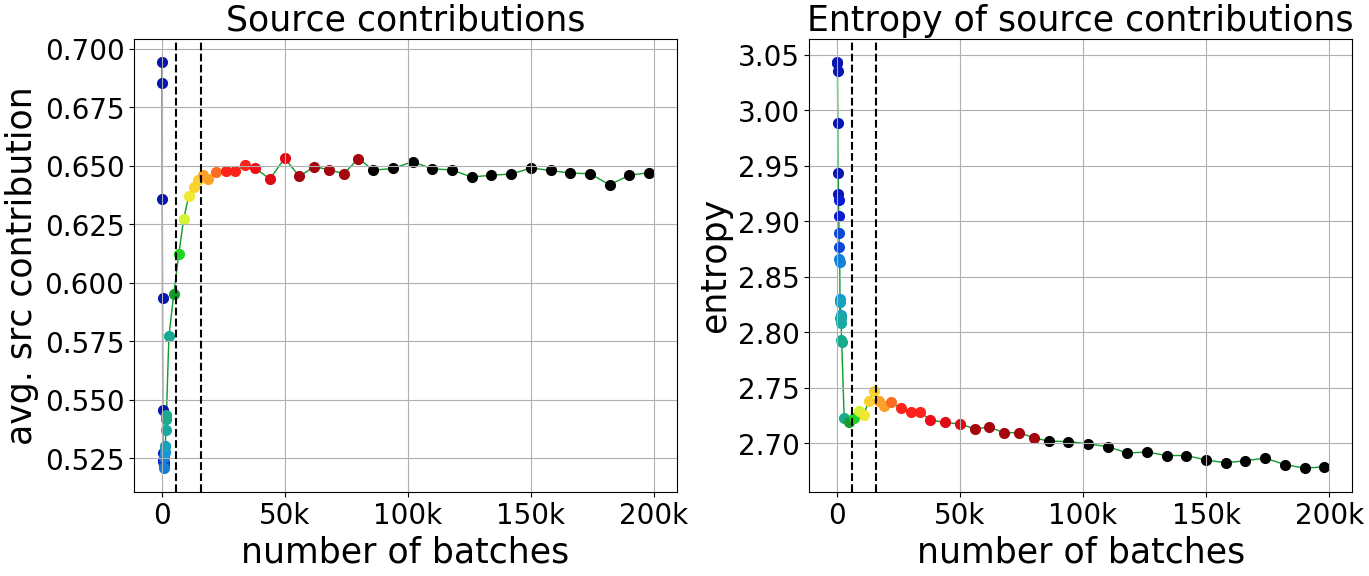}}
    \vspace{-2ex}
  \caption{Left: contribution of source, right: entropy of source contributions. En-De. Vertical lines separate the stages.}
  \vspace{-2ex}
    \label{fig:lrp_ende}
\end{figure}

\begin{figure}[t!]
    \centering
    {\includegraphics[scale=0.34]{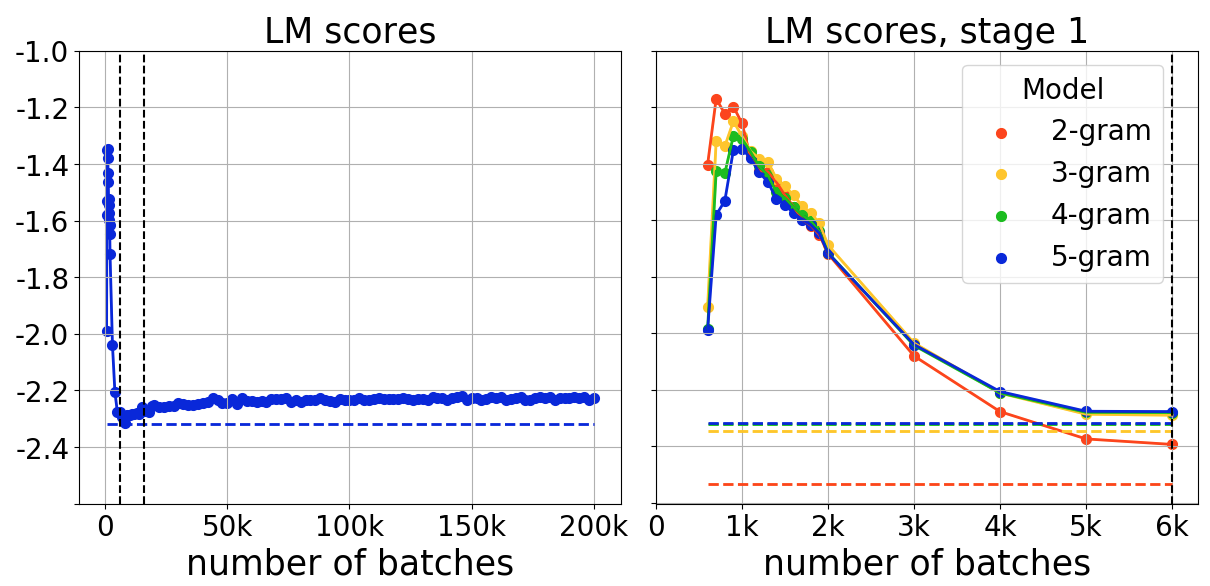}}
  \caption{KenLM scores. Left: 5-gram model, all training stages; right: different models, the first stage. Horizontal lines show the scores for the references. En-De.}
  \vspace{-2ex}
    \label{fig:lm_scores_ende}
\end{figure}

\begin{figure}[t!]
    \centering
    {\includegraphics[scale=0.3]{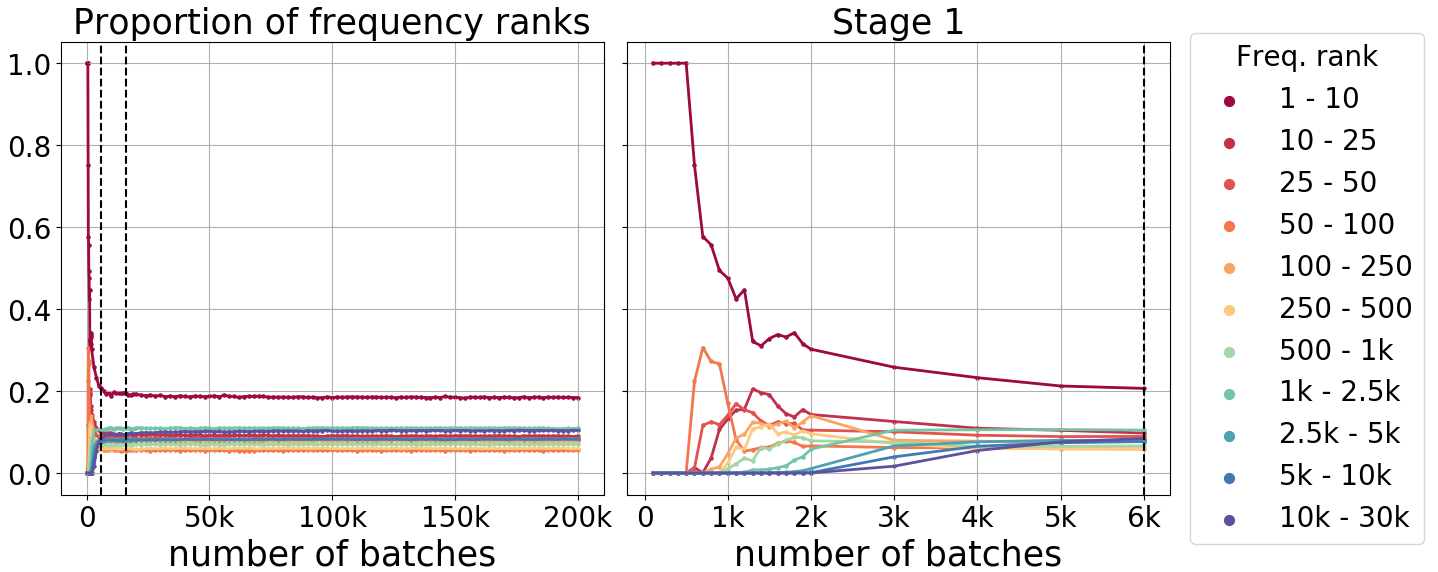}}
  \caption{Proportion of tokens of different frequency ranks in model translations. En-De.}
  \vspace{-2ex}
    \label{fig:freq_ranks_ende}
\end{figure}

\begin{figure}[t!]
    \centering
    \subfloat[]
    {\includegraphics[scale=0.19]{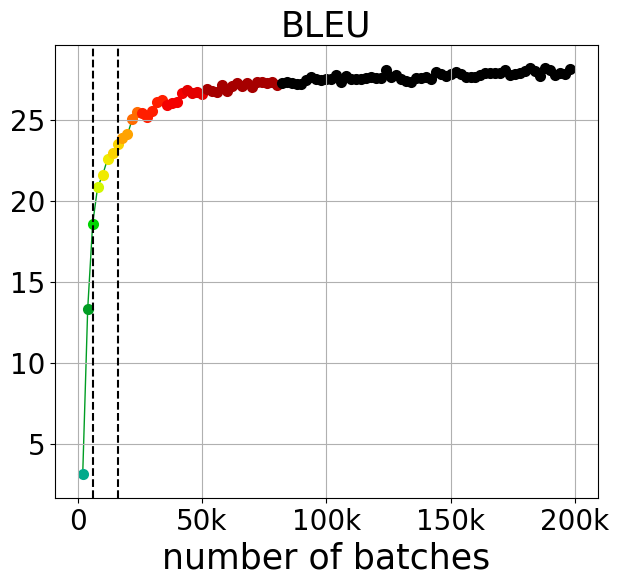}}
    \ 
    \subfloat[]
    {\includegraphics[scale=0.28]{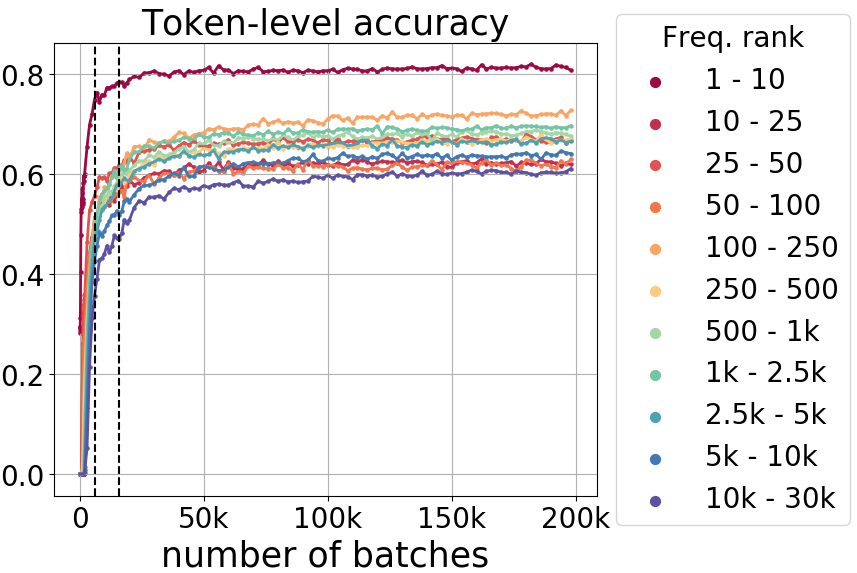}}
    \vspace{-1ex}
  \caption{(a) BLEU score; (b) token-level accuracy (the proportion of cases where the correct next token is the most probable choice). WMT En-De.}
  \vspace{-2ex}
    \label{fig:ende_bleu_acc}
\end{figure}

\begin{figure}[t!]
    \centering
    \subfloat[]
    {\ \ \includegraphics[scale=0.20]{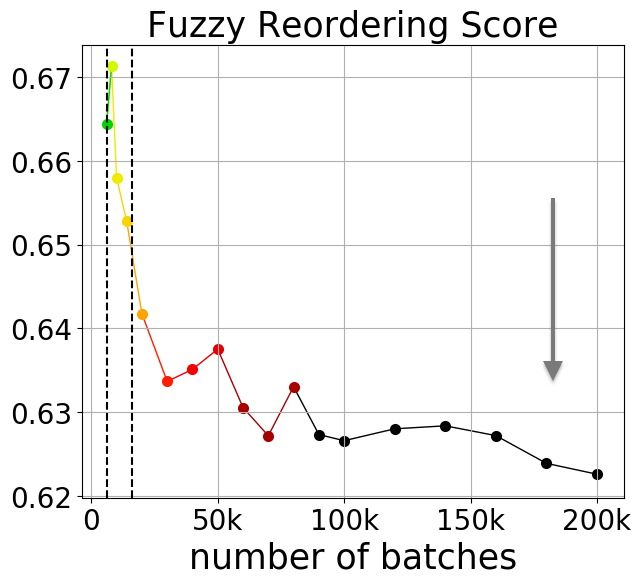}}
    \quad
    \subfloat[]
    {\includegraphics[scale=0.20]{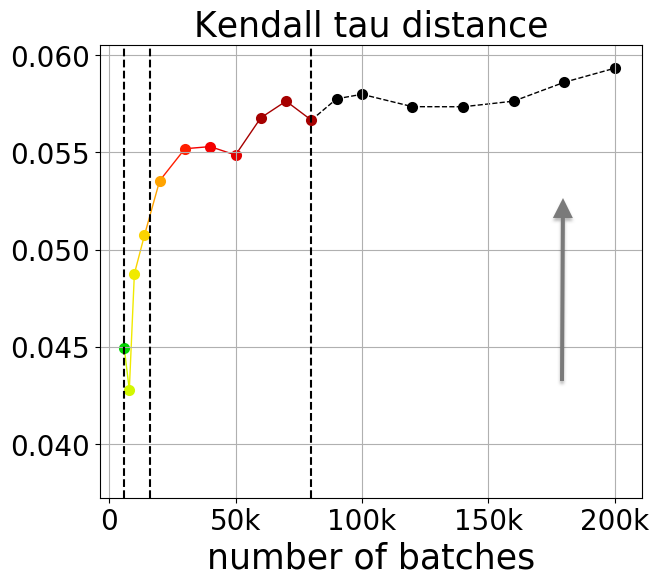}}
    \vspace{-1ex}
  \caption{(a) fuzzy reordering score (for references: 0.5), (b) Kendall tau distance (for references: 0.08); WMT En-De. The arrows point in the direction of less monotonic alignments (more complicated reorderings).}
  \vspace{-2ex}
    \label{fig:ende_frs}
\end{figure}

Figure~\ref{fig:lrp_ende} shows the abstract stages for En-De, Figures~\ref{fig:lm_scores_ende}-\ref{fig:ende_frs} provide the results from Section~\ref{sect:transformer_training_stages} for the other language pair (En-De).

\section{Other Models}

\begin{figure}[t!]
    \centering
    {\includegraphics[scale=0.23]{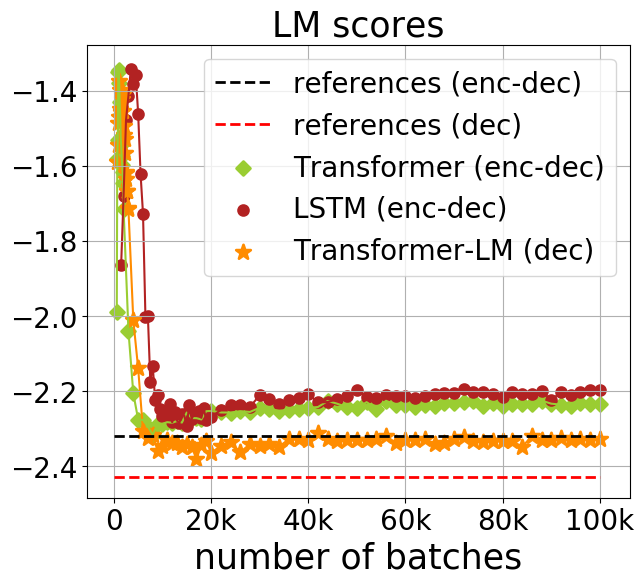}}
  \caption{Target-side LM scores (5-gram); En-De.}
    \label{fig:other_models_ende_appendix}
\end{figure}

\begin{figure}[t!]
    \centering
    \subfloat[]
    {\includegraphics[scale=0.23]{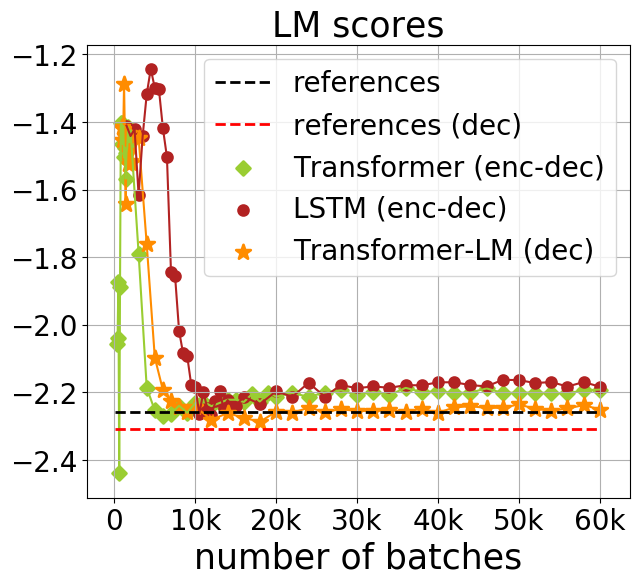}}
    \ \ 
    \subfloat[]
    {\includegraphics[scale=0.23]{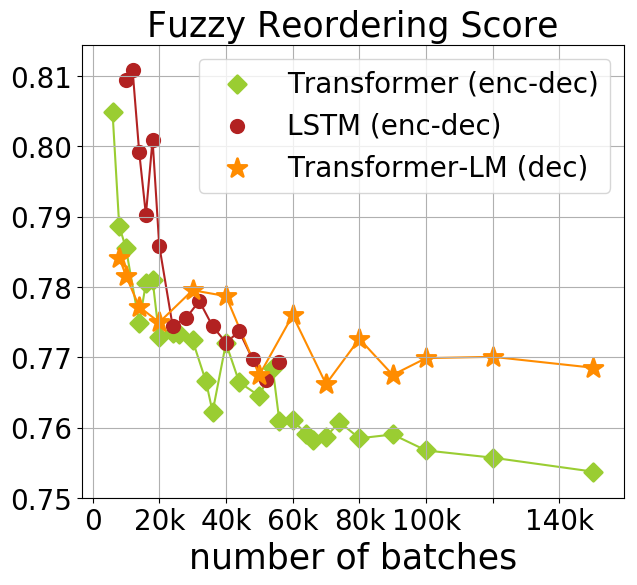}}
  \caption{(a) target-side LM scores, (b) fuzzy reordering score (for references: 0.6); WMT En-Ru.}
  \vspace{-2ex}
    \label{fig:other_models_enru}
\end{figure}

Figure~\ref{fig:other_models_ende_appendix} is a version of the Figure~\ref{fig:other_models_ende}a from the main text, but with the scores for all three models. Figure~\ref{fig:other_models_enru} provides corresponding results for the other language pair (En-Ru). Note that in Figure~\ref{fig:other_models_enru}b the reordering score for the LSTM model stops earlier: this is because the LSTM model converges earlier than other models.

\section{Practical Applications}

\subsection{Experimental Setting}

\paragraph{Model.} The model is the re-implemented by~\citet{Zhou2020Understanding} version of the vanilla NAT by~\citet{gu2018nonautoregressive}. Namely, instead of modeling fertility as described in the original paper, \citet{Zhou2020Understanding}  monotonically copy the encoder embeddings to the input of the decoder. We used the code released by~\citet{Zhou2020Understanding}.\footnote{ \url{https://github.com/pytorch/fairseq/tree/master/examples/nonautoregressive_translation}}

\paragraph{Training.} For all experiments, we follow the setting by~\citet{Zhou2020Understanding}. Note that in their work, training NAT models required 32 GPUs. In our setting, we ensured the same batch size by accumulating gradients for several batches (in \texttt{fairseq}, this is done using the \texttt{--update-freq} option).

\paragraph{NAT Inference.} Following previous work, for this vanilla NAT model we use a straight-forward decoding algorithm which simply picks the \texttt{argmax} at every position.